\documentclass[wcp]{jmlr}

\usepackage{array}
\usepackage{booktabs}
\usepackage{enumitem}
\usepackage{graphicx}
\usepackage{microtype}
\usepackage{tikz}
\usetikzlibrary{arrows.meta}
\newcommand{\method}{claim-gated audit}
\newcommand{\CBPS}{composition-bounded predictive support}
\newcommand{\labelProbe}{\texttt{probe\_\allowbreak positive}}
\newcommand{\labelAV}{\texttt{action\_value\_\allowbreak controlled\_\allowbreak positive}}
\newcommand{\labelShortcut}{\texttt{composition\_\allowbreak shortcut\_\allowbreak supported}}
\newcommand{\labelBelief}{\texttt{belief\_like\_\allowbreak supported}}
\newcommand{\labelCBPS}{\texttt{cbps\_\allowbreak supported}}
\newcommand{\scopeCBPS}{\texttt{diagnostic\_\allowbreak interpretation\_\allowbreak not\_\allowbreak ordered\_\allowbreak belief}}
\newcommand{\labelInsufficient}{\texttt{insufficient\_\allowbreak pairs}}
\newcommand{\labelInconclusive}{\texttt{inconclusive}}
\newcommand{\statusValidated}{\texttt{protocol\_\allowbreak validated}}
\newcommand{\statusApplicability}{\texttt{protocol\_\allowbreak applicability\_\allowbreak supported}}
\newcolumntype{L}[1]{>{\raggedright\arraybackslash}p{#1}}
\newcolumntype{C}[1]{>{\centering\arraybackslash}p{#1}}

\jmlryear{}
\jmlrworkshop{}
\title[Beyond Tracking or Shortcut]{Beyond Tracking or Shortcut: Composition-Bounded Predictive States in Poker Autoregressive Models}
\author[Quanhao Li and Qianyu Chen]{Quanhao Li and Qianyu Chen}
\editors{}

\begin{document}
\makeatletter
\let \@jmlrpages \@empty
\let \@jmlrproceedings \@empty
\let \@titlefoot \@empty
\let \@jmlrenddoc \relax
\def\@jmlrmaketitle{%
 {%
  \jmlrpretitle
  {\def\titletag##1{##1}\@title}%
  \jmlrposttitle
  \begin{center}
    {\normalsize\bfseries Quanhao Li$^{*,\ddagger}$ \quad Qianyu Chen$^{\dagger,\ddagger}$\par}
    \vspace{0.35em}
    {\small $^*$University of Massachusetts \quad $^\dagger$Abbey Park High School\par}
    \vspace{0.25em}
    {\small \texttt{quanhaoli@umass.edu} \quad \texttt{1924chenjor@gmail.com}\par}
    \vspace{0.6em}
    {\footnotesize $^\ddagger$Joint first-authorship.\par}
  \end{center}
  \vskip 0.2in
 }%
}
\def\ps@jmlrtps{%
  \let\@mkboth\@gobbletwo
  \def\@oddhead{}%
  \let\@evenhead\@oddhead
  \def\@oddfoot{}%
  \let\@evenfoot\@oddfoot
}
\makeatother
\maketitle

\begin{abstract}
Hidden-state probes often recover latent labels in imperfect-information sequence models, which, however, alone does not establish that a model maintains a posterior belief distribution over hidden states.
This paper aims to study this ambiguity in a no-range Limit Hold'em autoregressive model trained only on action and value targets, not an opponent's hand or range.
Opponent-range probes are positive after action/value controls in two of three seeds, and the behavior head predicts held-out actions about five accuracy points above a baseline using only the observable public history.
However, visible public betting composition explains more opponent-range signal than residual hidden states, suggesting that most of the recoverable information comes from betting summaries.
Action/value+composition baselines reach 16.5--16.7\% top-10 accuracy while composition-residual hidden probes fall to 11.4--12.2\%, and matched-composition comparisons are negative in every seed.
We call this evidence pattern composition-bounded predictive support: hidden states remain behavior-predictive and opponent-range correlated, but most recoverable range information is explained by visible betting composition rather than residual hidden-state structure.
This is a case-study claim about opponent-range representational evidence, not exact Bayes posterior tracking or a causal belief mechanism.
Synthetic control and oracle validations show that the same diagnostics accept posterior-sensitive states and reject raw composition states under matched controls.
Thus positive belief probes should be interpreted through targeted alternatives before being treated as evidence of belief tracking.
\end{abstract}


\section{Introduction}

Autoregressive models trained on imperfect-information histories must compress public observations into hidden states.
Because public histories are statistically correlated with hidden private variables, probes trained on frozen hidden states can often recover latent labels such as an opponent range.
That result is not self-interpreting.
The same probe score may reflect a posterior-sensitive state or a useful public-history compression whose statistics correlate with the hidden target.
Recoverability alone is therefore insufficient evidence for posterior-sensitive representations, since the same probe signal may arise from publicly observable betting patterns that happen to correlate with opponent range.

Therefore, the usual binary framing of either tracking or shortcutting is too coarse.
A positive probe is often read as belief tracking, while failed robustness checks are then described as mere shortcut behavior.
However, probe outcomes can also exist outside these extremes.
A behavior-trained sequence model may learn a predictive public-state representation: it is useful for action or value prediction and carries latent-correlated information, but the latent-label information is bounded by visible public-history composition and does not preserve opponent-range distinctions under matched-composition tests.
We refer to this pattern as composition-bounded predictive support: evidence that a representation is behaviorally useful and correlated with hidden variables, while remaining insufficient to support a stronger posterior-tracking interpretation.

We study this distinction as a case study in a no-range Limit Hold'em autoregressive (AR) model trained on behavior targets.
On a locked test split, the behavior head outperforms public-history lookup baselines on held-out action prediction across all three seeds.
A separate representation-to-behavior audit shows that frozen hidden states remain predictive of downstream targets, especially value, although much of this relationship is captured by input-only composition features.
Frozen hidden-state probes also recover nontrivial opponent-range signal: after action/value controls, residual top-10 range signal remains positive in two of three seeds, while the third remains inconclusive.
Taken together, these results reject the weakest interpretation that the hidden state contains no opponent-range information.

Stronger controls alter the interpretation.
Full visible-history composition features explain more range signals than residual hidden states, and residual representations fail to outperform action/value+composition priors within matched-composition buckets across all seeds.
A range label proxy gap analysis likewise finds no stable residual tracking once composition is controlled.
Accordingly, this case study provides evidence about opponent-range representations rather than exact Bayesian posterior tracking.
The best-supported interpretation is composition-bounded predictive support: the hidden state remains behaviorally useful and correlated with opponent range, but does not support a stronger posterior-tracking claim.

This refinement avoids two overstatements: the poker AR should not be described as having learned opponent-range belief, and the result should not be dismissed as a meaningless probe artifact.
Known-control, posterior/oracle, and non-poker learned-positive controls show that exact posterior states pass, raw composition states collapse, and the audit has a positive path.

Therefore, we summarize this paper's exclusive contributions as:
\begin{enumerate}[leftmargin=1.5em]
\item We introduce a claim-safe three-way interpretation for hidden-state belief probes: artifact, composition-bounded predictive support, and posterior-tracking support.
\item We show that a no-range Limit Hold'em AR model falls into the CBPS regime: its states are behavior-predictive and latent-correlated, but the latent signal is bounded by public-history composition.
\item We validate the distinction with known-control CBPS, posterior-state, oracle, non-poker latent-order rows, and a controlled behavior-trained positive benchmark.
\item We provide reproducible audit outputs that separate strict case verdicts, refined interpretations, oracle validation, and forbidden stronger claims.
\end{enumerate}

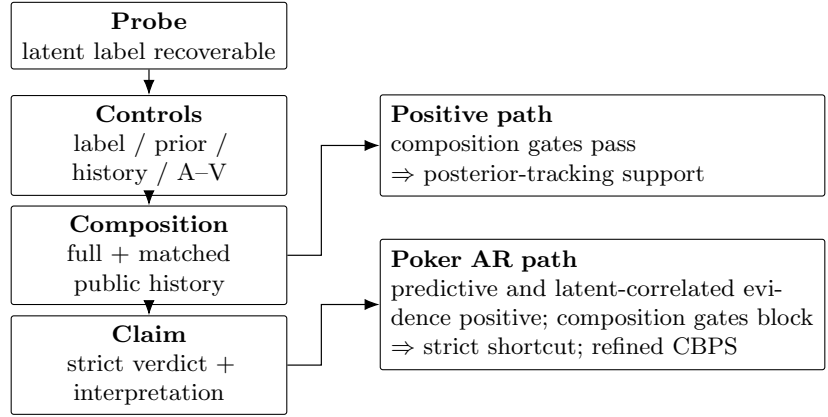
\begin{figure}[t]
\centering
\begin{tikzpicture}[
gate/.style={draw, rounded corners=1pt, align=center, text width=3.4cm, minimum height=0.88cm, inner sep=3.5pt, font=\footnotesize},
outcome/.style={draw, rounded corners=1pt, align=left, text width=5.6cm, minimum height=0.92cm, inner sep=4pt, font=\footnotesize},
arrow/.style={-{Latex[length=2mm,width=1.5mm]}, line width=0.45pt}
]
\node[gate] (probe) at (0,0) {\textbf{Probe}\\latent label recoverable};
\node[gate] (controls) at (0,-1.45) {\textbf{Controls}\\label / prior / history / A--V};
\node[gate] (composition) at (0,-2.90) {\textbf{Composition}\\full + matched public history};
\node[gate] (claim) at (0,-4.35) {\textbf{Claim}\\strict verdict + interpretation};
\draw[arrow] (probe.south) -- (controls.north);
\draw[arrow] (controls.south) -- (composition.north);
\draw[arrow] (composition.south) -- (claim.north);

\node[outcome] (positive) at (6.0,-1.45) {\textbf{Positive path}\\composition gates pass\\$\Rightarrow$ posterior-tracking support};
\node[outcome] (poker) at (6.0,-3.55) {\textbf{Poker AR path}\\predictive and latent-correlated evidence positive; composition gates block\\$\Rightarrow$ strict shortcut; refined CBPS};
\draw[arrow] (composition.east) -- ++(0.45,0) |- (positive.west);
\draw[arrow] (claim.east) -- ++(0.45,0) |- (poker.west);
\end{tikzpicture}
\caption{Claim-gated audit ladder. Probe recovery is routed through nuisance and composition gates before stronger representational claims are allowed. The poker AR follows the lower branch: positive predictive and latent-correlated evidence, but blocked composition gates.}
\label{fig:audit-ladder}
\end{figure}

Table~\ref{tab:mini-matrix} previews the evidence-to-claim mapping, and Figure~\ref{fig:audit-ladder} shows the claim-gated audit ladder.
The central paper contract is that protocol-validation rows do not upgrade the audited poker instance label, and the CBPS interpretation remains below an ordered posterior-tracking claim.

\begin{table}[t]
\centering
\caption{First-page claim preview. The poker case supports CBPS but not ordered posterior tracking; validation rows evaluate the audit rather than upgrading the poker case.}
\label{tab:mini-matrix}
\small
\begin{tabular}{@{}L{0.26\linewidth}L{0.39\linewidth}L{0.24\linewidth}@{}}
\toprule
Row & Gate outcome & Supported output \\
\midrule
Behavior-trained poker AR & predictive evidence is positive; weak range recovery is nontrivial, with action/value residuals positive in 2/3 seeds; full and matched composition gates block ordered posterior tracking & \CBPS{}; no belief upgrade \\
Known-control CBPS & constructed posterior, composition, and CBPS states are separated & middle case separated from alternatives \\
Posterior/oracle controls & ordered states and controlled learned positives pass, while shortcut states fail under matched composition & protocol validated \\
Poker stress tests & range-supervised, strong-teacher, and MLP rows do not clear strict composition gates & no belief upgrade \\
\bottomrule
\end{tabular}
\end{table}

\section{Related Work}

\noindent\textbf{Probing and claim control.}
Linear probes are widely used to inspect intermediate representations~\citep{alain-bengio-2016-linear-probes}.
Prior work shows that probe accuracy must be interpreted through controls rather than treated as direct evidence of a represented structure~\citep{hewitt-liang-2019-control,hewitt-manning-2019-structural,belinkov2022probing}.
Our setting adds a domain-specific confound: in imperfect-information behavior histories, visible public composition can be highly informative about the hidden target.
Thus the relevant question is not just whether the label is recoverable, but whether the recovery survives action/value, composition, and matched-composition alternatives.

\noindent\textbf{Belief states and imperfect information.}
In partially observable decision processes, a belief state is a posterior over latent state given observed history~\citep{kaelbling1998planning}.
Recurrent agents and world models use hidden state to integrate partial observations over time~\citep{hausknecht2015drqn,ha2018worldmodels}, and sequential-decision interpretability work studies how policies can be summarized or explained~\citep{bekkemoen24a,wang24c}.
This paper does not claim to identify a belief-state planner.
It asks when hidden activations justify the weaker or stronger representational claims that are often attached to belief probes.

\noindent\textbf{Reliability audits for interpretation.}
Interpretation methods can create a false sense of explanation when familiar signals are not checked against targeted alternatives.
Examples include saliency-map audits and robustness specifications~\citep{zhang25a,le25a}, attention interpretability analysis~\citep{pandey23a}, uncertainty-aware global explanations~\citep{gkolemis23a}, concept explanations for time series~\citep{obermair23a}, and robust counterfactual explanations~\citep{jiang24a}.
Our analogue is a claim-gated audit for hidden-state belief probes.
The CBPS interpretation is useful only because the gate structure keeps the allowed claim below posterior tracking.

\section{Interpretations and Audit Protocol}

Let $x_{1:t}$ be an observable history, $z_t$ a hidden target such as an opponent private-state class, and $h_t=f_\theta(x_{1:t})$ a frozen hidden activation.
A probe $q_\phi(z_t\mid h_t)$ tests recoverability.
The interpretation question asks which alternative explanations have been ruled out.

Table~\ref{tab:taxonomy} states the three interpretations used in this paper.
The middle row is the new framing.
It should not be confused with a mechanism claim: \CBPS{} means that the evidence supports a behavior-predictive, latent-correlated, composition-bounded public-state pattern under the audit.
It does not prove that a particular subspace causally mediates action selection.

\noindent\textbf{Definition 1 (Composition-bounded predictive support).}
For a frozen behavioral sequence model and latent target $Z$, \CBPS{} is the diagnostic status in which:
(i) held-out behavior prediction is positive beyond public-history lookup baselines;
(ii) frozen hidden states recover latent-correlated signal after label, prior, history, and action/value controls; and
(iii) full and matched public-history composition controls block an ordered posterior-tracking upgrade.
Thus CBPS names a positive but bounded evidence package: the state is behavior-predictive and latent-correlated, while the strongest surviving explanation for the stronger belief claim remains public-history composition.
CBPS is not assigned by the failure of posterior-tracking tests alone: it requires positive behavior-prediction evidence and positive latent-label recoverability before the composition block; otherwise the audit returns only an artifact, nuisance, shortcut, or inconclusive label.
It is not a causal mediation or posterior-tracking claim.

\begin{table}[t]
\centering
\caption{Three interpretations for positive hidden-state belief probes. The paper's no-range poker row supports the middle interpretation, not posterior tracking.}
\label{tab:taxonomy}
\scriptsize
\resizebox{\linewidth}{!}{%
\begin{tabular}{@{}L{0.19\linewidth}L{0.26\linewidth}L{0.24\linewidth}L{0.24\linewidth}@{}}
\toprule
Interpretation & What is recoverable & Diagnostic signature & Allowed claim \\
\midrule
Degenerate artifact & label prior, shuffled-label artifact, no-observation prior, or direct nuisance & weak controls explain the signal & probe result is not meaningful evidence \\
\CBPS{} & behavior-relevant public-history compression with latent-correlated signal & behavior prediction and weak range probes are positive, but full composition and matched-composition gates block posterior tracking & composition-bounded predictive support \\
Posterior tracking & order-sensitive posterior information about the hidden target & signal survives action/value, composition, and matched-composition posterior tests & belief-like evidence under specified gates \\
\bottomrule
\end{tabular}}
\end{table}

The \method{} maps evidence to admissible claims rather than presenting a dashboard of ablations.
A claim gate maps an evidence object to \texttt{pass}, \texttt{blocked}, \texttt{untested}, or \texttt{insufficient\_support}.
The case gates evaluate probe recoverability, label controls, no-observation priors, history controls, action/value nuisance, full visible composition, and matched-composition support.
The protocol returns a strict case label, an oracle/protocol status, and forbidden stronger claims.
This structure prevents the refined CBPS interpretation from silently replacing the stricter verdict.

\begin{algorithm}[t]
\footnotesize
\caption{Claim-Gated Audit with CBPS Interpretation}
\label{alg:audit}
\DontPrintSemicolon
\SetAlgoLined
\KwIn{$H,Z,X,A,V,C,B$; behavior report $P$; optional oracle suite $O$}
\KwOut{$y_{\mathrm{strict}}$, $y_{\mathrm{interp}}$, $s_{\mathrm{oracle}}$, $\mathcal{F}$}
\BlankLine
\textbf{Evaluate case gates.}\;
$G_1 \leftarrow \mathrm{probe}(H,Z)$\;
$G_2 \leftarrow \mathrm{label\_control}(H,\mathrm{shuffle}(Z))$\;
$G_3 \leftarrow \mathrm{prior\_control}(Z)$\;
$G_4 \leftarrow \mathrm{history\_control}(H,X,Z)$\;
$G_5 \leftarrow \mathrm{action\_value\_control}(H,A,V,Z)$\;
$G_6 \leftarrow \mathrm{comp\_control}(H,C,A,V,Z)$\;
$G_7 \leftarrow \mathrm{matched\_comp}(H,C,B,Z)$\;
\BlankLine
\textbf{Assign strict case verdict.}\;
\eIf{$G_7=\texttt{insufficient\_support}$}{
  $y_{\mathrm{strict}} \leftarrow$ \labelInsufficient\;
}{
  \eIf{$G_6=\texttt{blocked}$ \textbf{or} $G_7=\texttt{blocked}$}{
    $y_{\mathrm{strict}} \leftarrow$ \labelShortcut\;
  }{
    \eIf{$G_1,\ldots,G_7$ all pass}{
      $y_{\mathrm{strict}} \leftarrow$ \labelBelief\;
    }{
      \eIf{$G_5=\texttt{pass}$ and $G_6,G_7$ are untested}{
        $y_{\mathrm{strict}} \leftarrow$ \labelAV\;
      }{
        $y_{\mathrm{strict}} \leftarrow$ weaker or inconclusive label\;
      }
    }
  }
}
\BlankLine
\textbf{Refine interpretation without upgrading belief.}\;
\eIf{$P=\texttt{predictive\_supported}$ and $G_1,\ldots,G_5$ support latent signal and $G_6/G_7$ block posterior tracking}{
  $y_{\mathrm{interp}} \leftarrow$ \labelCBPS\;
}{
  $y_{\mathrm{interp}} \leftarrow y_{\mathrm{strict}}$\;
}
$s_{\mathrm{oracle}} \leftarrow \mathrm{ValidateOracle}(O)$\;
$\mathcal{F} \leftarrow \mathrm{ForbiddenClaims}(y_{\mathrm{strict}},y_{\mathrm{interp}},s_{\mathrm{oracle}})$\;
\Return{$y_{\mathrm{strict}},y_{\mathrm{interp}},s_{\mathrm{oracle}},\mathcal{F}$}\;
\end{algorithm}

Table~\ref{tab:gate-rules} states the gate semantics used throughout the paper.
A gate is not an ablation result by itself; it is a precondition for a stronger interpretation.
For G1--G4, the audit delta rule requires that the configured held-out delta is positive with the reported uncertainty source.
For G5--G7, the claim-critical comparisons use paired or bucket bootstrap intervals.
A \texttt{blocked} composition gate does not prove that the model lacks every form of belief-relevant information.
It blocks the stronger ordered-belief claim because a named observable alternative remains sufficient under the audit.

\begin{table}[t]
\centering
\caption{Claim gates. Deltas are evaluated on held-out data with locked layer discipline; CI denotes a 95\% paired or bucket bootstrap interval when available.}
\label{tab:gate-rules}
\scriptsize
\resizebox{\linewidth}{!}{%
\begin{tabular}{@{}L{0.12\linewidth}L{0.24\linewidth}L{0.28\linewidth}L{0.25\linewidth}@{}}
\toprule
Gate & Alternative explanation & Pass or block rule & Claim effect \\
\midrule
G1 Probe & no recoverable target signal & probe score exceeds reference with positive audit delta and reported uncertainty & permits \labelProbe{} \\
G2 Label & probe or label artifact & shuffled-label controls do not reproduce the signal under the audit delta rule & blocks label-artifact explanation \\
G3 Prior & class prior or no-observation shortcut & hidden-state signal exceeds no-observation prior with reported uncertainty & blocks fixed-prior explanation \\
G4 History & non-history or order-destroyed explanation & intact histories outperform destroyed histories under configured uncertainty & blocks non-history explanation \\
G5 A/V nuisance & action/value target confound & residual hidden probe has CI lower $>0$ against action/value baseline & permits \labelAV{} \\
G6 Composition & visible composition shortcut & blocked if composition-aware residual comparison CI upper $<0$ & assigns \labelShortcut{} \\
G7 Matched comp. & bucket-level composition confound & blocked if matched-bucket residual CI upper $<0$; insufficient if support is absent & assigns \labelShortcut{} or \labelInsufficient{} \\
O Oracle & protocol false-positive/false-negative behavior & ordered controls pass and shortcut controls are blocked & validates protocol only \\
\bottomrule
\end{tabular}}
\end{table}

Table~\ref{tab:protocol-comparison} makes the evidence-to-claim ladder explicit.
The CBPS interpretation is intentionally placed below posterior tracking: it requires positive predictive evidence and recoverable latent-correlated signal, while composition gates still block an ordered-belief upgrade.
This is the main safeguard against post-hoc renaming.
If the same evidence package lacks predictive support, the audit can still return a strict shortcut verdict, but it should not receive the CBPS interpretation.

\begin{table}[t]
\centering
\caption{Evidence packages and admissible claims. We separate the strict gate label from the refined CBPS interpretation so that bounded predictive evidence cannot be mistaken for posterior tracking.}
\label{tab:protocol-comparison}
\scriptsize
\resizebox{\linewidth}{!}{%
\begin{tabular}{@{}L{0.23\linewidth}L{0.25\linewidth}L{0.25\linewidth}L{0.19\linewidth}@{}}
\toprule
Evidence package & Alternatives still unresolved & Strongest admissible claim & Forbidden stronger claim \\
\midrule
Probe only & label artifact, prior, history destruction, action/value nuisance, composition shortcut & \labelProbe{} & belief-like support \\
Probe + label/prior/history controls & action/value nuisance and visible-composition shortcut & recoverable history-conditioned signal & belief-like support \\
Probe + action/value control & visible-composition shortcut remains & \labelAV{} & ordered belief tracking \\
Predictive behavior + weak latent signal + blocked composition gates & posterior tracking remains blocked but the state is not empty & strict \labelShortcut{}; refined \labelCBPS{} & posterior tracking or mechanism claim \\
All case gates pass & no named case alternative remains under the audit & \labelBelief{} & causal mechanism claim \\
Oracle/applicability controls & protocol behavior only; no case-label upgrade & \statusValidated{} or \statusApplicability{} & upgrading the audited case label \\
\bottomrule
\end{tabular}}
\end{table}

The audit configuration is fixed before interpreting the final label.
The main poker case uses locked layer \texttt{encoder\_layer\_2}, the predeclared full visible-history composition summary, 1,000 bootstrap resamples where applicable, and matched-composition support with reported buckets, pairs, and dropped fraction.
Oracle controls are evaluated separately from the audited case.
The supplementary early-gate manifest records the statistic, pass rule, uncertainty source, and artifact path for G1--G4.
The positive path is defined before the poker downgrade: exact posterior-state, oracle posterior, and belief-auxiliary latent-order representations can pass matched-composition gates, while composition shortcut controls collapse.
Those rows validate protocol behavior; they do not upgrade the audited poker AR.

\section{Known Controls Separate CBPS from Belief}

A CBPS label would be empty if the audit could only distinguish posterior states from raw composition states.
We therefore use a known-control environment with three constructed states.
The posterior state carries the exact posterior and should support posterior tracking.
The composition state carries only visible composition and should collapse under matched composition.
The CBPS state is predictive and latent-correlated but bounded: it carries a nontrivial representation difference under matched composition while its posterior-gap slope is not stably positive.

Table~\ref{tab:known-cbps} shows that the three labels are separated.
The CBPS row has a large behavior-prediction delta against $C_{\min}$, a positive label-control delta, and nonzero matched representation JS, but the posterior-gap slope interval crosses zero.
This validates CBPS as a diagnostic label.
It does not by itself claim that the poker AR has CBPS; the poker row still needs behavior and boundedness evidence.

\begin{table}[t]
\centering
\caption{Known-control CBPS validation. The audit distinguishes exact posterior tracking, raw composition, and a composition-bounded predictive state.}
\label{tab:known-cbps}
\scriptsize
\resizebox{\linewidth}{!}{%
\begin{tabular}{@{}L{0.18\linewidth}L{0.21\linewidth}rrrr@{}}
\toprule
State & Output & Action delta vs $C_{\min}$ & Label delta & Repr. JS & Posterior-gap slope CI \\
\midrule
Posterior state & posterior tracking & -0.061523 & 0.447903 & 0.429045 & [1.000000, 1.000000] \\
Composition state & composition shortcut & 0.000000 & 0.035244 & 0.000000 & [0.000000, 0.000000] \\
CBPS state & \labelCBPS{} & 0.597168 & 0.017603 & 0.036409 & [-0.013733, 0.003230] \\
\bottomrule
\end{tabular}}
\end{table}

The action-delta column validates the predictive requirement for CBPS, whereas the posterior-state row is judged by posterior-sensitive representation alignment.
Thus a posterior state can validate the posterior-tracking path without satisfying the CBPS-specific predictive criterion.
The CBPS state is also not the composition state: its matched representation JS is nonzero while the composition state collapses.

\section{Poker: Predictive and Range-Correlated}

The audited poker model is a four-layer AR Transformer with hidden size 96 and four attention heads.
It is trained on action and value behavior targets with policy loss weight 1.0, value loss weight 0.3, and opponent-hand loss weight 0.0.
For each seed, 50,000 teacher-policy hands are split by hand id into deterministic 60/20/20 train, validation, and test partitions.
Opponent-range labels are used only after training for probes.
The main rows use locked layer \texttt{encoder\_layer\_2}; best-layer curves are descriptive only.
The probe target has 169 opponent-range classes, so uniform top-10 chance is $10/169=5.9\%$.
The relevant baselines are stronger empirical nuisances: no-observation prior, action/value baseline, composition-only baseline, and action/value+composition baseline.
The action/value baseline trains the same linear classifier on public action and value features.
The visible composition summary contains length, capped length, action counts, action proportions, action-presence indicators, first-action and last-action indicators, and position summaries for bet, call, check, and fold.
It is predeclared and excludes hidden cards, opponent-range labels, teacher logits, future actions, and future outcomes.
The held-out teacher-policy action mix is stable across no-range seeds: bet about 0.26, call 0.11, check 0.48, fold 0.15, with about 2.55 prior actions per decision.
This makes composition a natural nuisance for poker belief claims, not an arbitrary post-hoc feature.

\noindent\textbf{Behavior prediction.}
The first requirement for CBPS is predictive support.
The behavior head predicts held-out action targets above public-history lookup baselines across all three seeds.
Against the full visible-composition lookup baseline, action-accuracy deltas are about five points with tight confidence intervals (Table~\ref{tab:predictive-link}).
This is not causal mediation evidence, and it does not identify which hidden subspace drives behavior.
It supports the narrower claim that the no-range AR row is behavior-predictive beyond public-history lookup baselines.

\begin{table}[t]
\centering
\caption{Full-test predictive link. The behavior head predicts action targets beyond public-history lookup baselines; this supports the predictive part of CBPS but not causal mediation.}
\label{tab:predictive-link}
\small
\begin{tabular}{@{}lrrrr@{}}
\toprule
Seed & Test points & Action acc. & Delta vs $C_{\mathrm{full}}$ & 95\% CI \\
\midrule
k & 49,480 & 0.837086 & 0.050728 & [0.048302, 0.053416] \\
l & 49,258 & 0.839661 & 0.053149 & [0.050307, 0.055849] \\
m & 49,132 & 0.837112 & 0.052715 & [0.050171, 0.055422] \\
\bottomrule
\end{tabular}
\end{table}

\noindent\textbf{Representation-to-behavior link.}
The representation-to-behavior audit asks whether frozen hidden representations themselves predict behavior targets when read out by independent linear probes.
Against input-only $C_{\mathrm{full}}$ linear features, hidden probes have mean action-accuracy delta +0.006609 with seed interval [-0.000407, 0.010489], and value-MSE improvement +0.318198 with seed interval [0.225321, 0.395974].
After residualizing hidden features against the same $C_{\mathrm{full}}$ inputs, action and value deltas are negative (-0.222154 and -0.484330).
The resulting gate is \texttt{composition\_bounded\_behavior\_link}: it strengthens the predictive part of CBPS, but remains correlational and composition-bounded rather than causal mediation evidence.

\noindent\textbf{Range-correlated signal.}
The action/value baseline is around 13.4--13.6\% top-10 accuracy.
Residual hidden-state probes after action/value control reach 14.0--14.6\%.
Seed \texttt{k} and seed \texttt{m} have positive lower confidence bounds; seed \texttt{l} is inconclusive.
Thus the hidden state is not empty with respect to opponent range, but this evidence is still below the standard needed for an ordered posterior-tracking claim.

\begin{table}[t]
\centering
\caption{Locked-layer range signal after action/value controls. These rows establish latent-correlated signal under weaker gates, not posterior tracking.}
\label{tab:weak-probes}
\small
\begin{tabular}{@{}lrrrrL{0.22\linewidth}@{}}
\toprule
Seed & Residual top-10 & A/V baseline & Delta & 95\% CI & Claim after A/V \\
\midrule
k & 0.146403 & 0.134600 & 0.013216 & [0.008472, 0.018230] & \labelAV{} \\
l & 0.140546 & 0.134049 & 0.001844 & [-0.002755, 0.006694] & \labelInconclusive{} \\
m & 0.142453 & 0.135859 & 0.006417 & [0.001473, 0.011366] & \labelAV{} \\
\bottomrule
\end{tabular}
\end{table}

\section{The Poker Signal Is Composition-Bounded}

\begin{figure}[t]
\centering
\includegraphics[width=\linewidth]{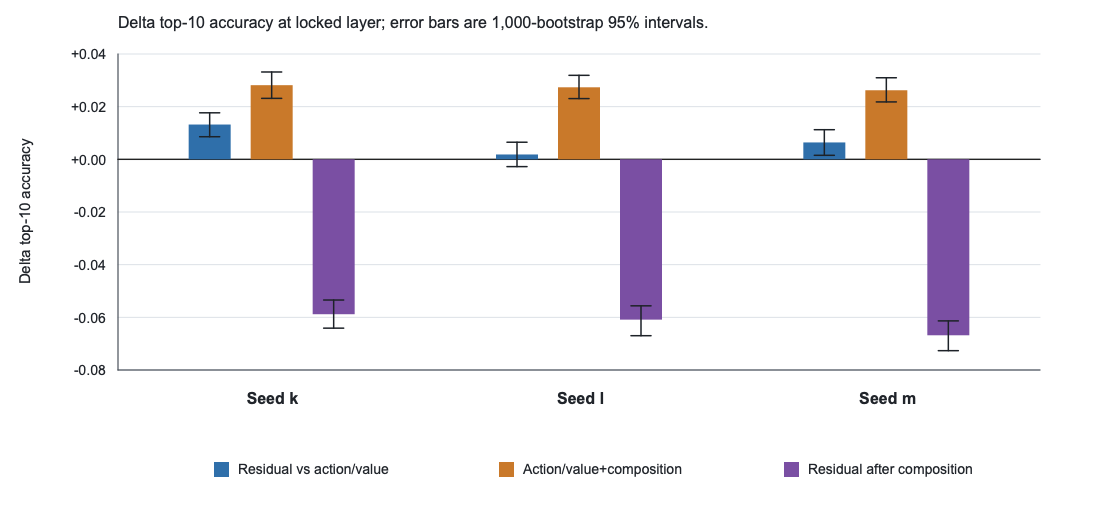}
\caption{Visible-composition baselines dominate residual hidden-state range signal. This is the core boundedness evidence: action/value controls are not enough for posterior tracking when full public-history composition remains predictive.}
\label{fig:composition-per-seed}
\end{figure}

The decisive question is whether the range-correlated signal exceeds visible public-history composition.
It does not.
Action/value+composition baselines reach 16.5--16.7\% top-10 accuracy, while the composition-residual hidden probe falls to 11.4--12.2\%.
Figure~\ref{fig:composition-per-seed} shows that composition-aware nuisance baselines dominate residual hidden-state probes across seeds.

Matched-composition evaluation rules out a weaker objection: perhaps the composition baseline is only globally stronger, while residual hidden states still carry order-sensitive opponent-range evidence inside comparable histories.
Table~\ref{tab:composition} answers that objection.
Within matched buckets, residual hidden states do not beat action/value+composition priors in any seed, with confidence intervals below zero and 146 matched buckets per seed.
The full artifact records matched support details beyond the bucket count; the main text reports the claim-critical per-seed deltas.
The strict gate verdict remains \labelShortcut{}.

\begin{table}[t]
\centering
\caption{Matched-composition evidence. `CBPS interp.' denotes the refined interpretation after combining predictive support with composition-boundedness; the strict gate output remains shortcut.}
\label{tab:composition}
\scriptsize
\resizebox{\linewidth}{!}{%
\begin{tabular}{@{}lrrrrL{0.22\linewidth}@{}}
\toprule
Seed & Hidden residual & A/V+comp. prior & Delta & 95\% CI & Output \\
\midrule
k & 0.122413 & 0.166390 & -0.063352 & [-0.080361, -0.047150] & shortcut; CBPS \\
l & 0.121666 & 0.165455 & -0.062623 & [-0.081007, -0.045442] & shortcut; CBPS \\
m & 0.114141 & 0.166511 & -0.075168 & [-0.093948, -0.056334] & shortcut; CBPS \\
\bottomrule
\end{tabular}}
\end{table}

\begin{figure}[t]
\centering
\includegraphics[width=0.92\linewidth]{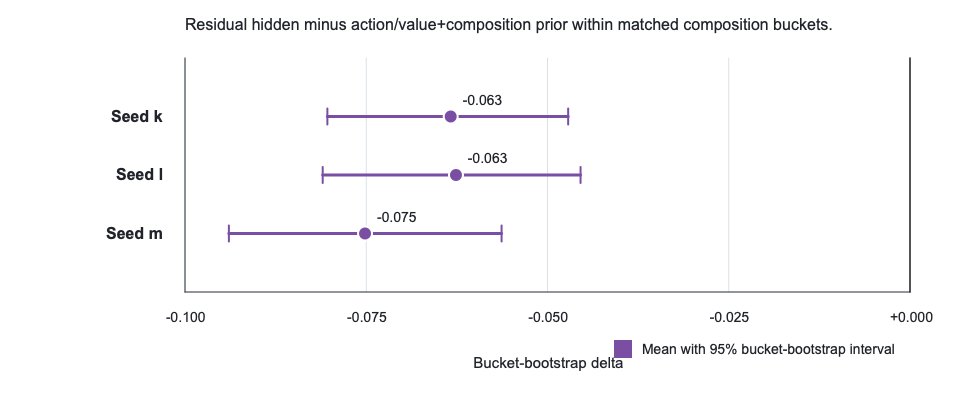}
\caption{Matched-composition deltas remain negative across seeds. The hidden residual does not preserve range evidence beyond matched public composition.}
\label{fig:matched-composition}
\end{figure}

As a same-domain gate-acceptance check, the poker oracle range-representation addendum replaces the audited AR hidden state with an oracle range representation under the same matched-composition support.
The gate then accepts ordered evidence in all three seeds, with minimum lower delta 0.833489.
This validates that the poker matched-composition gate can accept range-representation evidence when the representation actually contains it under same-domain support.
It does not upgrade the no-range AR hidden-state label.

The boundedness audit adds two safeguards.
First, the composition ladder shows that weaker summaries $C_{\min}$ and $C_{\mathrm{mid}}$ are visible proxy confounds, while the full visible-history summary $C_{\mathrm{full}}$ is the decisive residual-block gate.
Second, a matched range-label proxy analysis finds no stable positive residual tracking: seed confidence intervals are [-0.017343, 0.009236], [-0.026485, 0.000776], and [-0.008733, 0.023498].
This proxy analysis is not exact poker posterior evidence.
It is boundedness support, and the manuscript keeps the claim at \labelCBPS{} rather than posterior tracking.

\begin{table}[t]
\centering
\caption{Composition-feature sensitivity. Weak composition summaries expose visible proxy confounds; the predeclared full visible-history summary is the residual-blocking gate used for the strict verdict.}
\label{tab:composition-sensitivity}
\scriptsize
\resizebox{\linewidth}{!}{%
\begin{tabular}{@{}L{0.17\linewidth}L{0.34\linewidth}rL{0.31\linewidth}@{}}
\toprule
Summary & What it controls & Minimum lower bound & Claim effect \\
\midrule
$C_{\min}$ & coarse visible counts and legal/public summaries & 0.002114 & proxy confound remains visible \\
$C_{\mathrm{mid}}$ & richer public-history composition without the full declared set & 0.021482 & proxy confound remains visible \\
$C_{\mathrm{full}}$ & full declared visible-history composition used by G6/G7 & -0.093948 & decisive residual block \\
\bottomrule
\end{tabular}}
\end{table}

This sensitivity result is important for the CBPS interpretation.
If a very weak composition summary were the only blocker, the downgrade could look like an artifact of an arbitrary hand-engineered control.
Instead, weaker summaries show that public-history proxies are informative, and the full declared summary supplies the actual blocking evidence.
The conclusion is therefore not that any composition feature explains the hidden state.
It is that the audit's visible public-history composition is a sufficient alternative explanation for the stronger posterior-tracking claim.

\section{Claim-Boundary Ledger}

We keep several stress rows as a claim-boundary ledger rather than as main positive evidence.
Range-supervised poker, MLP probe capacity, strong-teacher students, and linear range-probe subspace ablation each test a natural upgrade path.
None clears the strict composition-sensitive gates required for posterior tracking.
Their role is therefore narrow: they prevent stronger mechanism or belief claims from being inferred from probe recovery, while the main CBPS claim rests on the positive predictive evidence, recoverable latent-correlated signal, and the decisive composition bound.
The full per-seed rows are reported in the supplement.

\section{Posterior Controls Show the Positive Path}

A strict audit should not only reject.
We validate the distinction in settings where the correct representational structure is known or controlled.
For ordered controls, the representation is the posterior or is trained to carry the posterior.
For shortcut controls, the representation is a function of visible composition.
When histories match on visible composition but differ in posterior, ordered controls should preserve representation divergence while shortcut controls should collapse.
In notation, an ordered control uses $r_{\mathrm{order}}(x_{1:t})=p(z\mid x_{1:t})$, while a shortcut control uses $r_{\mathrm{short}}(x_{1:t})=g(C(x_{1:t}))$.
When $C(x_i)=C(x_j)$ but $p(z\mid x_i)\neq p(z\mid x_j)$, the ordered control should retain representation divergence and the shortcut control should not.

Table~\ref{tab:validation} summarizes the validation and applicability rows.
The explicit posterior-state RNN has posterior JS 0.462898 and representation JS 0.462898, with lower confidence bound 0.446727, while the composition-state RNN collapses to 0.000000.
The latent-order benchmark is non-poker: each report has 2,048 sequences, four hidden types, length five, 111 matched buckets, and 4,096 matched pairs.
Across three reports, the aggregate has 333 matched buckets and 12,288 matched pairs.
Oracle posterior and belief-auxiliary GRU controls pass 3/3 reports; composition shortcut controls fail 3/3.
The behavior-only GRU has high representation divergence but remains inconclusive, showing why divergence alone is not enough for a belief claim.
This boundary row is useful because it separates order-sensitive variation from claim-sufficient belief evidence.
An order-rich exact-posterior benchmark supplies the stronger learned-positive check: a behavior-only GRU trained only on posterior-sensitive behavior targets passes matched-composition diagnostic gates, and projection/removal of its posterior-correlated subspace changes behavior outputs more than same-rank random subspaces across three seeds.
These rows validate the audit's positive path without upgrading the Limit Hold'em poker AR verdict.

\begin{table}[t]
\centering
\caption{Posterior and learned-positive validations. These rows show that the audit can accept posterior-sensitive states and controlled behavior-trained positives without upgrading the poker case label.}
\label{tab:validation}
\scriptsize
\resizebox{\linewidth}{!}{%
\begin{tabular}{@{}L{0.24\linewidth}rrrL{0.20\linewidth}L{0.25\linewidth}@{}}
\toprule
Control or benchmark row & Buckets & Pairs & Representation JS & Output & Gate reason \\
\midrule
Explicit posterior-state RNN & -- & 1,024 & 0.462898 & posterior state passes & matched posterior representation passes \\
Explicit composition-state RNN & -- & 1,024 & 0.000000 & \labelShortcut{} & matched representation collapses \\
Latent-order oracle posterior state & 333 & 12,288 & 0.470037 & belief-like applicability 3/3 & posterior oracle passes \\
Latent-order belief-aux GRU hidden probe & 333 & 12,288 & 0.487425 & belief-like applicability 3/3 & supervised ordered positive passes \\
Latent-order composition shortcut states & 333 & 12,288 & 0.000000 & shortcut 3/3 & shortcut state collapses \\
Latent-order action/value-only GRU & 333 & 12,288 & 0.497270 & inconclusive 3/3 & high divergence, full gate not cleared \\
Order-rich behavior-only GRU & 1,476 & 30,000 & 0.341688 & learned posterior supported & behavior-trained learned positive passes \\
Order-rich projection stress & -- & 55,986 & -- & projection supported 3/3 & posterior-correlated subspace affects behavior outputs \\
\bottomrule
\end{tabular}}
\end{table}

\begin{figure}[t]
\centering
\includegraphics[width=\linewidth]{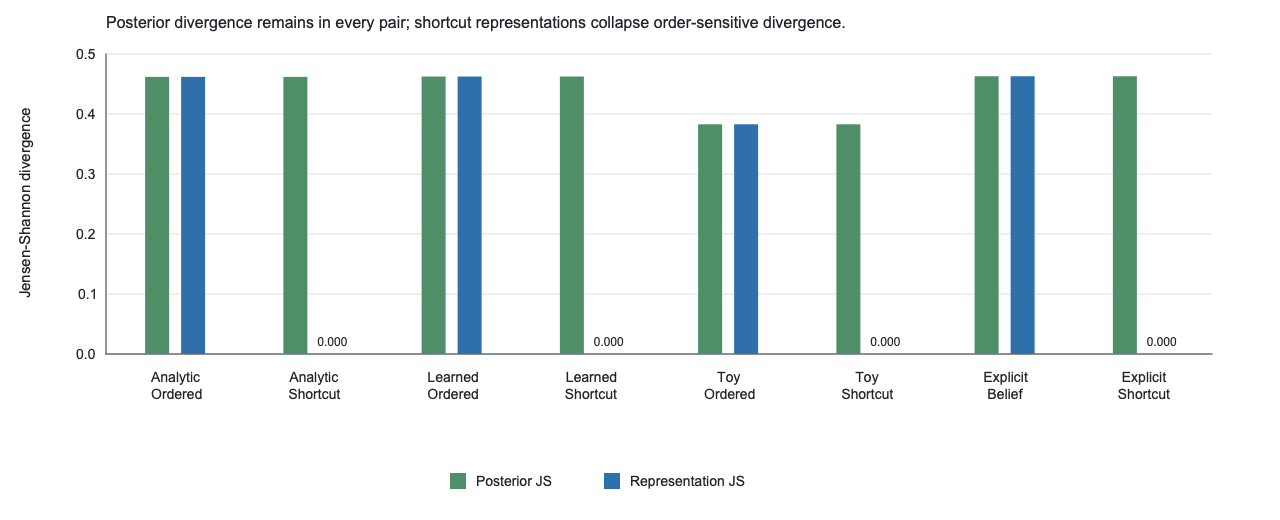}
\caption{Posterior-sensitive controls preserve matched-composition representation divergence; composition controls collapse. These rows calibrate the interpretation and do not upgrade the poker label.}
\label{fig:oracle-validation}
\end{figure}

Figure~\ref{fig:oracle-validation} is the methodological counterpart to the poker downgrade.
The ordered controls preserve representation divergence when the posterior changes under same-composition counterfactuals.
The shortcut controls keep nonzero posterior divergence but collapse representation divergence to zero.
The audit can therefore accept ordered posterior representations and reject composition shortcuts when the correct answer is known.
The poker result is not a consequence of a rejection-only protocol; the audited poker case simply does not satisfy the positive path.
The controlled learned-positive benchmark strengthens the audit validation but also preserves the paper's claim discipline: random-source patching is not specific, so we report robust projection-level behavior relevance rather than full patch-specific causal belief tracking.

\section{Discussion}

The main lesson is that positive belief probes recover public-history information whose interpretation must be diagnosed.
For the no-range poker AR row, the hidden state is not empty: it supports behavior prediction and carries recoverable opponent-range signal under weaker gates.
Yet the range signal is composition-bounded.
Full public-history composition explains more than the residual hidden state, and matched-composition tests block the ordered posterior-tracking claim.

CBPS is useful because it names the supported middle interpretation without erasing the strict verdict.
The strict gate label remains \labelShortcut{}.
The refined interpretation is \labelCBPS{} only with the scope \scopeCBPS{}.
This means the model has a behavior-predictive public-state pattern whose latent correlation and behavior-link evidence are bounded by visible composition.
An input-level order-intervention stress test further bounds the shortcut reading: when earlier betting actions are permuted while preserving counts and the last action, hidden states and outputs move across all three seeds.
Because these synthetic perturbations are not guaranteed to be legal poker trajectories, this shows order sensitivity but does not upgrade the claim to legal posterior tracking.
It does not mean that the model lacks all belief-like information, and it does not identify a causal mechanism.

The controlled positives are equally important.
Posterior-state, oracle, latent-order, and controlled learned-positive rows show that the audit can accept posterior-sensitive representations when they are present.
In the learned-positive benchmark, a behavior-only GRU passes matched-composition diagnostic gates and has robust projection-level behavior relevance across three seeds.
The known-control CBPS row shows that the middle label is distinguishable from both raw composition and posterior tracking.
Together, these controls make the poker result more informative than a negative probe audit.

The practical implication is that belief-probing papers should report not only whether a latent label is decodable, but also which public-history explanations remain sufficient.
CBPS is useful because it preserves a positive statement about the learned state while preventing an unsupported posterior-tracking upgrade.

\section{Reproducibility}

The replication package reports both the strict verdict and the refined interpretation.
The source manifest records:
\begin{center}
\small
\begin{tabular}{@{}ll@{}}
\toprule
Strict verdict & \labelShortcut{} \\
CBPS status & \labelCBPS{} \\
CBPS behavior link & \texttt{composition\_bounded\_behavior\_link} \\
CBPS scope & \scopeCBPS{} \\
\bottomrule
\end{tabular}
\end{center}
This prevents a presentation error in which the safer CBPS language silently replaces the stricter gate outcome.
The audit operationalizes the claim gates rather than presenting a post-hoc script bundle.
Given an archive root, the regeneration path produces diagnostic reports, strict and refined claim labels, paper tables, figure data, source manifests, and SHA-256 checksums.
The manifest links each reported row to its source artifact, including the synthetic CBPS control, poker predictive-link report, representation-to-behavior report, composition-boundedness report, poker oracle gate-acceptance report, controlled learned-positive benchmark, composition-sensitivity report, early-gate threshold manifest, and capacity or stress-test reports.
The complete audit records validation status for posterior-state, synthetic, toy, learned-positive, range-supervised, capacity, and latent-order rows without using those rows to upgrade the audited poker case.
The paper figures are generated from JSON-backed figure packs with source checksums.
Dense claim matrices, frozen-LM textual stress rows, strong-teacher rows, full per-seed stress tables, teacher action-composition rows, and full early-gate manifests are supplementary material.
This split is intentional: the main paper shows the evidence path and decisive gate outcomes, while the supplement documents the full audit ledger and auxiliary checks.

\section{Limitations}

The main behavioral CBPS row is one Limit Hold'em AR family with three seeds.
Controlled posterior, known-CBPS, latent-order, learned-positive, and POMDP addenda improve interpretation, but they do not prove that all behavior-trained imperfect-information AR models learn CBPS.
The order-rich learned-positive benchmark is not the Limit Hold'em case, and its intervention evidence is projection-level rather than full patch-specific causal belief tracking.
The poker boundedness proxy uses range-label gaps rather than exact posterior JS.
The predictive-link and representation-to-behavior results are correlational and do not establish causal mediation from a representation subspace to behavior.
The range-probe subspace ablation stress test is negative under the tested linear row-space intervention, so it further cautions against reading probe recoverability as behavior mediation.
The MLP row is inconclusive, not a proof that higher probe capacity cannot reveal ordered evidence.
Mechanistic claims would require exact-posterior poker abstraction, activation patching, stronger targeted interventions, or other identification assumptions.

\section{Conclusion}

Positive hidden-state belief probes should be treated as the start of an interpretation problem.
In the audited no-range Limit Hold'em AR row, behavior prediction is positive and opponent-range signal is recoverable, but the signal is bounded by visible public-history composition and does not support ordered posterior tracking.
The best-supported interpretation is composition-bounded predictive support, not belief tracking and not an empty artifact.
Belief probing should therefore move from asking only whether a latent label is recoverable to asking what kind of public-history state makes it recoverable.

\bibliography{jorjor_refs}

\end{document}